\documentclass[11pt]{article}
\usepackage{coling2016}
\usepackage{times}
\usepackage{latexsym}
\usepackage{color}
\usepackage{booktabs}
\usepackage{multirow}
\usepackage{amsmath}
\usepackage{amssymb}
\usepackage{url}
\usepackage{graphicx}
\usepackage{tipa}
\usepackage{algorithm}
\usepackage{algpseudocode}

\usepackage{subcaption}

\newcommand{\hl}[1]{\textbf{#1}}

\newcommand*{\thead}[1]{\multicolumn{1}{c}{\bfseries #1}}

\title{Faster decoding for subword level Phrase-based SMT between related languages}

\author{Anoop Kunchukuttan, Pushpak Bhattacharyya \\
Center For Indian Language Technology, \\
Department of Computer Science \& Engineering \\
Indian Institute of Technology Bombay \\
{\tt \{anoopk,pb\}@cse.iitb.ac.in}
}

\date{}

\begin{document}

\maketitle

\begin{abstract}	

A common and effective way to train translation systems between related languages is to consider sub-word level basic units. However, this increases the length of the sentences resulting in increased decoding time. The increase in length is also impacted by the specific choice of data format for representing the sentences as subwords. In a phrase-based SMT framework, we investigate different choices of \textit{decoder parameters} as well as \textit{data format} and their impact on decoding time and translation accuracy. We suggest best options for these settings that significantly improve decoding time with little impact on the translation accuracy. 
 
\end{abstract}

\blfootnote{
    \hspace{-0.65cm}
    This work is licenced under a Creative Commons
    Attribution 4.0 International License.
    License details:\\
    \url{http://creativecommons.org/licenses/by/4.0/}
} 

\section{Introduction}

\textit{Related languages} are those that exhibit lexical and structural similarities on account of sharing a \textbf{common ancestry} or being in \textbf{contact for a long period of time} \cite{bhattacharyya2016statistical}. Examples of languages related by common ancestry are Slavic and Indo-Aryan languages. Prolonged contact leads to convergence of linguistic properties even if the languages are not related by ancestry and could lead to the formation of \textit{linguistic areas} \cite{thomason2000linguistic}. Examples of such linguistic areas are the Indian subcontinent \cite{emeneau1956india}, Balkan \cite{trubetzkoy1928proposition} and Standard Average European \cite{haspelmath2001european} linguistic areas. Both forms of language relatedness lead to related languages sharing vocabulary and structural features. 

There is substantial government, commercial and cultural communication among people speaking related languages (Europe, India and South-East Asia being prominent examples and linguistic regions in Africa possibly in the future). As these regions  integrate more closely and move to a digital society, translation between \textit{related} languages is becoming an important requirement. In addition, translation to/from related languages to a \textit{lingua franca} like English is also very important. However, in spite of significant communication between people speaking related languages, most of these languages have few parallel corpora resources. It is therefore important to leverage the relatedness of these languages to build good-quality statistical machine translation (SMT) systems given the lack of parallel corpora. 

Modelling the lexical similarity among related languages is the key to building good-quality SMT systems with limited parallel corpora. \textbf{Lexical  similarity} implies that the languages share many words with the similar form (spelling/pronunciation) and meaning \textit{e.g.} \texttt{blindness} is \texttt{andhapana} in Hindi, \texttt{aandhaLepaNaa} in Marathi.  These words could be cognates, lateral borrowings or loan words from other languages. 

Sub-word level transformations are an effective way for translation of such shared words. Using subwords as basic units of translation has been shown to be effective in improving translation quality with limited parallel corpora. Subword units like character \cite{vilar2007can,tiedemann2009character}, character n-gram \cite{tiedemann2013analyzing} and orthographic syllables \cite{kunchukuttan2016orthographic} have been explored and have been shown to improve translation quality to varying degrees. 

However, the use of subword units increases the sentence length. This increases the training, tuning and decoding time for phrase-based SMT systems by an order of magnitude. This makes experimentation costly and time-consuming and impedes faster feedback which is important for machine translation research. Higher decoding time also makes deployment of MT systems based on subword units impractical. 

In this work, we systematically study the choice of data format for representing sentences and various decoder parameters which affect decoding time. Our studies show that the use of cube-pruning during tuning as well as testing with a lower value of the stack pop limit parameter improves decoding time substantially with minimal change in translation quality. 

The rest of the paper is organized as follows. Section \ref{sec:factors_decode_time} discusses the factors that affect decoding time which have been studied in this paper. Section \ref{sec:exp_setup} discusses our experimental setup. Section \ref{sec:results_decoder_params} discusses the results of our experiments with decoder parameters.  Section \ref{sec:results_corpus_format} discusses the results of our experiments with corpus formats. Section \ref{sec:related_work} discusses prior work related to optimizing decoders for phrase-based SMT. Section \ref{sec:conclusion} concludes the paper. 

\section{Factors affecting decoding time}
\label{sec:factors_decode_time}

This section describes the factors affecting the decoding time that have been studied in this paper. 

\subsection{Unit of translation}

The decoding time for a sentence is proportional to length of the sentence (in terms of the basic units). Use of subword units will obviously result in increased sentence length. Various units have been proposed for translation (character, character n-gram, orthographic syllable, morpheme, etc.). We analysed the average length of the input sentence on four language pairs (Hindi-Malayalam, Malayalam-Hindi, Bengali-Hindi, Telugu-Malayalam) on the ILCI corpus \cite{jha2012ilci}. The average length of an input sentence for character-level representation is 7 times of the word-level input, while it is 4 times the word-level input for orthographic syllable level representation. So, the decoding time will increase substantially. 

\subsection{Format for sentence representation}

\begin{table}[t]
\centering
\begin{tabular}{ll}
\toprule
Original & \scriptsize{\texttt{this is an example of data formats for segmentation}} \\
Subword units & \scriptsize{\texttt{thi s i s a n e xa m p le o f da ta fo rma t s fo r se gme n ta tio n}} \\
\midrule
Internal Marker & \scriptsize{\texttt{thi\_ s i\_ s a\_ n e\_ xa\_ m\_ p\_ le o\_ f da\_ ta fo\_ rma\_ t\_ s fo\_ r se\_ gme\_ n\_ ta\_ tio\_ n}} \\
Boundary Marker & \scriptsize{\texttt{thi s\_ i s\_ a n\_ e xa m p le\_ o f\_ da ta\_ fo rma t s\_ fo r\_ se gme n ta tio n\_}} \\
Space Marker & \scriptsize{\texttt{thi s \_ i s \_ a n \_ e xa m p le \_ o f \_ da ta \_ fo rma t s \_ fo r \_ se gme n ta tio n}} \\
\bottomrule 
\end{tabular}
\caption{Formats for sentence representation with subword units (example of orthographic syllables)}
\label{tbl:data_format}
\end{table}

The length of the sentence to be decoded also depends on how the subword units are represented. We compare three popular formats for representation, which are illustrated in Table \ref{tbl:data_format}: 

\begin{itemize}
\item \textbf{Boundary Marker}: The subword at the boundary of a word is augmented with a marker character. There is one boundary subword, either the first or the last chosen as per convention. Such a representation has been used in previous work, mostly related to morpheme level representation.  

\item \textbf{Internal Marker}: Every subword internal to the word is augmented with a marker character. This representation has been used rarely, one example being the Byte Code Encoding representation used by University of Edinburgh's Neural Machine Translation system \cite{williams2016edinburgh,sennrich2015neural}. 

\item \textbf{Space Marker}: The subword units are not altered, but inter-word boundary is represented by a space marker. Most work on translation between related languages has used this format. 
\end{itemize}

For boundary and internal markers, the addition of the marker character does not change the sentence length, but can create two representations for some subwords (corresponding to internal and boundary positions), thus introducing some data sparsity. On the other hand, space marker doubles the sentence length (in terms in words), but each subword has a unique representation. 

\subsection{Decoder Parameters}

Given the basic unit and the data format, some important decoder parameters used to control the search space can affect decoding time. The decoder is essentially a search algorithm, and we investigated important settings related to two search algorithms used in the \textit{Moses} SMT system: (i) stack decoding, (ii) cube-pruning \cite{chiang2007hierarchical}.  We investigated the following parameters: 

\begin{itemize}
\item \textbf{Beam Size}: This parameter controls the size of beam which maintains the best partial translation hypotheses generated at any point during stack decoding. 
\item \textbf{Table Limit}: Every source phrase in the phrase table can have multiple translation options. This parameter controls how many of these options are considered during stack decoding. 
\item \textbf{Cube Pruning Pop Limit}: In the case of cube pruning, the parameter limits the number of hypotheses created for each stack .
\end{itemize}

Having a lower value for each of these parameters reduces the search space, thus reducing the decoding time. However, reducing the search space may increase search errors and decrease translation quality. Our work studies this time-quality trade-off.

\section{Experimental Setup}
\label{sec:exp_setup}

In this section, we describe the language pairs and datasets used, the details of our experiments and evaluation methodology.

\subsection{Languages and Dataset} 
We experimented with four language pairs (Bengali-Hindi, Malayalam-Hindi, Hindi-Malayalam and Telugu-Malayalam). Telugu and Malayalam belong to the Dravidian language family which are agglutinative. Bengali and Hindi are Indo-Aryan languages with a relatively poor morphology. The language pairs chosen cover  different combinations of morphological complexity between source and target languages. 

We used the multilingual ILCI corpus for our experiments \cite{jha2012ilci}, consisting of sentences from tourism and health domains. The data split is as follows -- \textit{training: 44,777, tuning: 1000, test: 500} sentences. 

\subsection{System details} 
As an example of subword level representation unit, we have studied the orthographic syllable (OS) \cite{kunchukuttan2016orthographic} in our experiments. 
The OS is a linguistically motivated, variable length unit of translation, which consists of one or more consonants followed by a vowel (a C$^+$V unit). 
But our methodology is not specific to any subword unit. Hence, the results and observations should hold for other subword units also. 
We used the \textit{Indic NLP Library}\footnote{\url{http://anoopkunchukuttan.github.io/indic_nlp_library}} for orthographic syllabification.  

Phrase-based SMT systems were trained with OS as the basic unit. We used the \textit{Moses} system \cite{koehn2007moses},  with \textit{mgiza}\footnote{\url{https://github.com/moses-smt/mgiza}} for alignment, the \textit{grow-diag-final-and} heuristic for symmetrization of word alignments, and Batch MIRA \cite{cherry2012batch} for tuning. Since data sparsity is a lesser concern due to small vocabulary size and higher order n-grams are generally trained for translation using sub-word units \cite{vilar2007can}, we trained 10-gram language models. The language model was trained on the training split of the target language corpus.  

The PBSMT systems were trained and decoded on a server with Intel Xeon processors (2.5 GHz) and 256 GB RAM. 

\subsection{Evaluation}

We use BLEU \cite{papineni2002bleu} for evaluating translation accuracy. We use the sum of user and system time minus the time for loading the phrase table (all reported by Moses) to determine the time taken for decoding the test set. 

\section{Effect of decoder parameters}
\label{sec:results_decoder_params}

\begin{table}[t]
\centering
{\small
\begin{tabular}{lrrrrrrrr}
\toprule
 & \multicolumn{4}{c}{\textbf{Translation Accuracy}} & \multicolumn{4}{c}{\textbf{Relative Decoding Time}} \\
\cmidrule(lr){2-5}
\cmidrule(lr){6-9}
 & \thead{ben-hin} & \thead{hin-mal} & \thead{mal-hin} & \thead{tel-mal} & \thead{ben-hin} & \thead{hin-mal} & \thead{mal-hin} & \thead{tel-mal} \\
\midrule
default (stack,  & \multirow{2}{*}{33.10} & \multirow{2}{*}{11.68} & \multirow{2}{*}{19.86} & \multirow{2}{*}{9.39} & \multirow{2}{*}{46.44} & \multirow{2}{*}{65.98} & \multirow{2}{*}{87.98} & \multirow{2}{*}{76.68} \\
tl=20,ss=100) & & & & & & & &  \\
\midrule
\multicolumn{9}{l}{\underline{\textit{Stack}}}\\
tl=10 & 32.84 & 11.24 & 19.21 & 9.47 & 35.49 & 48.37 & 67.32 & 80.51 \\
tl=5 & 32.54 & 11.01 & 18.39 & 9.29 & 15.05 & 21.46 & 30.60 & 41.52 \\
\midrule
ss=50 & 33.10 & 11.69 & 19.89 & 9.36 & 17.33 & 25.81 & 35.76 & 43.45 \\
ss=10 & 33.04 & 11.52 & 19.51 & 9.38 & 4.49 & 7.32 & 10.18 & 11.75 \\
\, +tuning & 32.83 & 11.01 & 19.57 & 9.23 & 5.24 & 8.85 & 11.60 & 9.31 \\
\midrule
\multicolumn{9}{l}{\underline{\textit{Cube Pruning}}}\\
pl=1000 & 33.05 & 11.47 & 19.66 & 9.42 & 5.67 & 9.29 & 12.38 & 17.85 \\
\, +tuning & 33.12 & 11.3 & 19.77 & 9.35 & 7.68 & 13.06 & 15.18 & 14.56\\
pl=100 & 32.86 & 10.97 & 18.74 & 9.15 & 2.00 & 4.22 & 5.41 & 5.29 \\
pl=10 & 31.93 & 9.42 & 15.26 & 8.5 & 1.51 & 3.64 & 4.57 & 3.84\\
\midrule
\midrule
\textit{Word-level}  & 31.62 & 9.67 & 15.69 & 7.54 & 100.56 ms & 65.12 ms & 50.72 ms & 42.4 ms\\
\bottomrule
\end{tabular}
}
\caption{Translation accuracy and Relative decoding time for orthographic syllable level translation using different decoding methods and parameters. Relative decoding time is indicated as a multiple of word-level decoding time. The following methods \& parameters in \textit{Moses} have been experimented with: (i) \textit{normal stack decoding} - vary \texttt{ss}: stack-size, \texttt{tt}: table-limit; (ii) \textit{cube pruning}: vary \texttt{pl}:cube-pruning-pop-limit. \texttt{+tuning} indicates that the decoder settings mentioned on previous row were used for tuning too. Translation accuracy and decode time per sentence for word-level decoding (in milliseconds) is shown on the last line for comparison.}
\label{tbl:decoder_params_accuracy}
\end{table}

We observed that the decoding time for OS-level models is approximately 70 times of the word-level model. 
This explosion in the decoding time makes translation highly compute intensive and difficult to perform in real-time. It also makes tuning MT systems very slow since tuning typically requires multiple decoding runs over the tuning set. Hence, we experimented with some heuristics to speed up decoding. 

For normal stack decoding, two decoder parameters which impact the decode time are: (1) \textit{beam size} of the hypothesis stack, and (2) \textit{table-limit}: the number of translation options for each source phrase considered by the decoder. Since the vocabulary of the OS-level model is far less than that of the word-level model, we hypothesize that lower values for these parameters can reduce the decoding time without significantly affecting the accuracy. Table \ref{tbl:decoder_params_accuracy} shows the results of varying these parameters.  We can see that with a beam size of 10, the decoding time is now about 9 times that of word-level decoding. This is a 7x improvement in decoding time over the default parameters, while the translation accuracy drops by less than 1\%.  If a beam size of 10 is used while decoding too, the drop in translation accuracy is larger (2.5\%). Using this beam size during decoding also slightly reduces the translation accuracy. On the other hand, reducing the table-limit significantly reduces the translation accuracy, while resulting in lesser gains in decoding time. 

We also experimented with \textit{cube-pruning} \cite{chiang2007hierarchical}, a faster decoding method first proposed for use with hierarchical PBSMT. The decoding time is controlled by the \texttt{pop-limit} parameter in the Moses implementation of cube-pruning. With a pop-limit of 1000, the decoding time is about 12 times that of word-level decoding. The drop in translation accuracy is about 1\% with a 6x improvement over default stack decoding, even when the model is tuned with a pop-limit of 1000. Using this pop-limit during tuning also hardly impacts the translation accuracy. However, lower values of pop-limit reduce the translation accuracy. 

While our experiments primarily concentrated on OS as the unit of translation, we also compared the performance of stack decoding and cube pruning for character lavel models. The results are shown in Table \ref{tbl:char_decoder_params_accuracy}. We see that character level models are 4-5 times slower than OS level models and hundreds of time slower than word level models with the default stack decoding. In the case of character based models also, the use of cube pruning (with \texttt{pop-limit}=1000) substantially speeds up decoding (20x speedup) with only a small drop in BLEU score. 

To summarize, we show that reducing the beam size for stack decoding as well as using cube pruning help to improve decoding speed significantly, with only a marginal drop in translation accuracy. Using cube-pruning while tuning only marginally impacts translation accuracy. 

\begin{table}[t]
\centering
{\small
\begin{tabular}{lrrrrrrrr}
\toprule
 & \multicolumn{4}{c}{\textbf{Translation Accuracy}} & \multicolumn{4}{c}{\textbf{Relative Decoding Time}} \\
\cmidrule(lr){2-5}
\cmidrule(lr){6-9}
 & \thead{ben-hin} & \thead{hin-mal} & \thead{mal-hin} & \thead{tel-mal} & \thead{ben-hin} & \thead{hin-mal} & \thead{mal-hin} & \thead{tel-mal} \\
\midrule
default (stack,  & \multirow{2}{*}{27.29} & \multirow{2}{*}{6.72} & \multirow{2}{*}{12.69} & \multirow{2}{*}{6.06} & \multirow{2}{*}{206.98} & \multirow{2}{*}{391.00} & \multirow{2}{*}{471.96} & \multirow{2}{*}{561.00} \\
tl=20,ss=100) & & & & & & & &  \\
\midrule
Cube Pruning & \multirow{2}{*}{26.98} & \multirow{2}{*}{6.57} & \multirow{2}{*}{11.94} & \multirow{2}{*}{5.99} & \multirow{2}{*}{10.23} & \multirow{2}{*}{19.57} & \multirow{2}{*}{24.59} & \multirow{2}{*}{26.20} \\
pl=1000 &  &  &  &  &  &  &  &  \\
\midrule
\midrule
\textit{Word-level}  & 31.62 & 9.67 & 15.69 & 7.54 & 100.56 ms & 65.12 ms & 50.72 ms & 42.4 ms\\
\bottomrule
\end{tabular}
}
\caption{Translation accuracy and Relative decoding time for character level translation using different decoding methods and parameters. Relative decoding time is indicated as a multiple of word-level decoding time. Translation accuracy and decode time per sentence for word-level decoding (in milliseconds) is shown on the last line for comparison.}
\label{tbl:char_decoder_params_accuracy}
\end{table}

\section{Effect of corpus format}
\label{sec:results_corpus_format}

For these experiments, we used the following decoder parameters: cube-pruning with cube-pruning-pop-limit=1000 for tuning as well as testing. Table \ref{tbl:formats_accuracy} shows the results of our experiments with different corpus formats.

The internal boundary marker format has a lower translation accuracy compared to the other two formats whose translation accuracies are comparable. In terms of decoding time, no single format is better than the others across all languages. Hence, it is recommended to use the space or boundary marker format for phrase-based SMT systems. Neural MT systems based on encoder decoder architectures, particularly without attention mechanism, are more sensitive to sentence length, so we presume that the boundary marker format may be more appropriate. 

\begin{table}[t]
\centering
{\small
\begin{tabular}{lrrrrrrrr}
\toprule
 & \multicolumn{4}{c}{\textbf{Translation Accuracy}} & \multicolumn{4}{c}{\textbf{Relative Decoding Time}} \\
\cmidrule(lr){2-5}
\cmidrule(lr){6-9}
 & \thead{ben-hin} & \thead{hin-mal} & \thead{mal-hin} & \thead{tel-mal} & \thead{ben-hin} & \thead{hin-mal} & \thead{mal-hin} & \thead{tel-mal} \\
\midrule
Boundary Marker & 32.83 & \hl{12.00} & \hl{20.88} & 9.02 & \hl{7.44} & 11.80 & 17.08 & 18.98 \\
Internal Marker & 30.10 & 10.53 & 19.08 & 7.53 & 7.82 & \hl{10.81} & \hl{14.43} & 17.06 \\
Space Marker & \hl{33.12} & 11.30 & 19.77 & \hl{9.35} & 7.68 & 13.06 & 15.18 & \hl{14.56} \\
\midrule
\midrule
\textit{Word-level}  & 31.62 & 9.67 & 15.69 & 7.54 & 100.56 ms & 65.12 ms & 50.72 ms & 42.4 ms \\
\bottomrule
\end{tabular}
}
\caption{Translation accuracy and Relative decoding time for orthographic syllable level translation using different data formats. Relative decoding time is indicated as a multiple of word-level decoding time. Translation accuracy and decode time per sentence (in milliseconds) for word-level decoding is shown on the last line for comparison.}
\label{tbl:formats_accuracy}
\end{table}

\section{Related Work}
\label{sec:related_work}

It has been recognized in the past literature on translation between related languages that the increased length of subword level translation is challenge for training as well as decoding \cite{vilar2007can}. Aligning long sentences is computationally expensive, hence most work has concentrated on corpora with short sentences (\textit{e.g.} OPUS \cite{tiedemann2009news}) \cite{tiedemann2009character,nakov2012combining,tiedemann2012character}. To make alignment feasible,  \newcite{vilar2007can} used the phrase table learnt from word-level alignment, which will have shorter parallel segments, as parallel corpus for training subword-level models. \newcite{tiedemann2013analyzing} also investigated  reducing the size of the phrase table by pruning, which actually improved translation quality for character level models. The authors have not reported the decoding speed, but it is possible that pruning may also improve decoding speed since fewer hypothesis may have to be looked up in the phrase table, and smaller phrase tables can be loaded into memory.    

There has been a lot of work looking at optimizing specific components of SMT decoders in a general setting. \newcite{hoang2016fast} provide a good overview of various approaches to optimizing decoders. Some of the prominent efforts include efficient language models \cite{heafield2011kenlm}, lazy loading \cite{zens2007efficient}, phrase-table design \cite{junczys2012space}, multi-core environment issues \cite{fernandez2016boosting}, efficient memory allocation \cite{hoang2016fast}, alternative stack configurations \cite{hoang2016fast} and alternative decoding algorithms like cube pruning \cite{chiang2007hierarchical}. 

In this work, we have investigated stack decoding configurations and cube pruning as a way of optimizing decoder performance for the translation between related languages (with subword units and monotone decoding). Prior work on comparing stack decoding and cube-pruning has been limited to word-level models \cite{huang2007forest,heafield2014faster}. 

\section{Conclusion and Future Work}
\label{sec:conclusion}

We systematically study the choice of data format for representing subword units in sentences and various decoder parameters which affect decoding time in a phrase-based SMT setting. Our studies (using OS and character as basic units) show that the \textbf{use of cube-pruning during tuning as well as testing with a lower value of the stack pop limit parameter} improves decoding time substantially with minimal change in translation quality. Two data formats, the space marker and the boundary marker, perform roughly equivalently in terms of translation accuracy as well as decoding time. Since the tuning step contains a decoder in the loop, these settings also reduce the tuning time. We plan to investigate reduction of the time required for alignment.

\section*{Acknowledgments}
We thank the Technology Development for Indian Languages (TDIL) Programme and the Department of Electronics \& Information Technology, Govt. of India for their support. We also thank the anonymous reviewers for their feedback. 

\bibliography{vardial3_2016_rel_lang_decode}
\bibliographystyle{acl}

\end{document}